\documentclass[runningheads]{llncs}
\usepackage[T1]{fontenc}
\usepackage{graphicx}
\usepackage{soul}
\usepackage{url}
\usepackage[utf8]{inputenc}
\usepackage{graphicx}
\usepackage{amsmath}
\usepackage{booktabs}
\usepackage[switch]{lineno}
\usepackage{todonotes}
\usepackage{multirow}
\usepackage{amsmath} 
\usepackage{bm}
\usepackage{booktabs}
\usepackage{multirow}
\usepackage{adjustbox}
\usepackage{xurl} 
\usepackage{seqsplit}
\usepackage{amsfonts}
\usepackage{amsmath, amssymb}

\usepackage[colorlinks=true, linkcolor=blue, urlcolor=blue, citecolor=blue]{hyperref}

\newcommand{\spara}[1]{\paragraph{#1}}
\newcommand{\dataparagraph}[1]{%
  \par\vspace{4pt}%
  \noindent{\normalfont\normalsize\itshape #1}\enspace%
}
\usepackage{marvosym}
\usepackage{xcolor}

\newcommand{\corrauth}{\textsuperscript{(\scalebox{1.25}{\Letter})}}

\begin{document}
\title{RTSKG: Building a Rail Transit Station Knowledge Graph Dataset}
%
%
\author{Shutong Zhu\inst{1} \and
Tianxing Wu\inst{1,2}\corrauth \and
Runfeng Liu\inst{1} \and Yuang Gu\inst{1} \and Xuan He\inst{1} \and Yuan Zhu\inst{3}}
\authorrunning{S. Zhu et al.}
%

\institute{School of Computer Science and Engineering, Southeast University, China
 \email{\{shutong\_zhu,tianxingwu\}@seu.edu.cn} \and
Key Laboratory of New Generation Artificial Intelligence Technology and Its Interdisciplinary Applications (Southeast University), Ministry of Education,  China \and
School of Architecture, Southeast University, China
}

%
\maketitle              

\setcounter{footnote}{0}

\begin{abstract}
Rail transit systems play a vital role in urban mobility and economic development. As key components of such systems, rail transit stations function as critical transport hubs that enhance urban accessibility and stimulate development in surrounding areas. City-level rail transit station related tasks (e.g., ridership prediction) require large-scale urban data, but current studies often neglect complex interactions among various urban entities in terms of data organization. In this paper, to address the above issue, we build a \underline{\textbf{R}}ail \underline{\textbf{T}}ransit \underline{\textbf{S}}tation \underline{\textbf{K}}nowledge \underline{\textbf{G}}raph (\textbf{RTSKG}) dataset which explicitly models the spatial and semantic interactions among different kinds of urban entities, to benefit city-level rail transit station related tasks. RTSKG integrates heterogeneous urban entities, such as rail transit stations, road segments, and points of interest, with a specially designed unified schema, and is accessible as Linked Data at \href{https://w3id.org/rtskg/}{https://w3id.org/rtskg/}. Evaluations on station-area store recommendation and knowledge-enhanced ridership prediction demonstrate the effectiveness of RTSKG, highlighting its potential to support city-level rail transit station analysis.

\end{abstract}

\section{Introduction}
Rail transit system is a local rail system providing passenger service within and around urban or suburban areas. As a main form of public transport, rail transit system is characterized by high capacity, high speed, and a high level of safety
\cite{LIN2022104509}, while also offering advantages in 
efficiency and punctuality over other modes of transportation like conventional bus systems~\cite{app13053022,DING2021125847,https://doi.org/10.1155/2022/8349173}. It plays a vital role in urban mobility and economic development since it can decrease congestion-related delay and help the development along transit corridors~\cite{weisbrod2009economic,to_centrality_2015}.

\begin{figure}[t]
  \centering
  \includegraphics[width=0.76\textwidth]{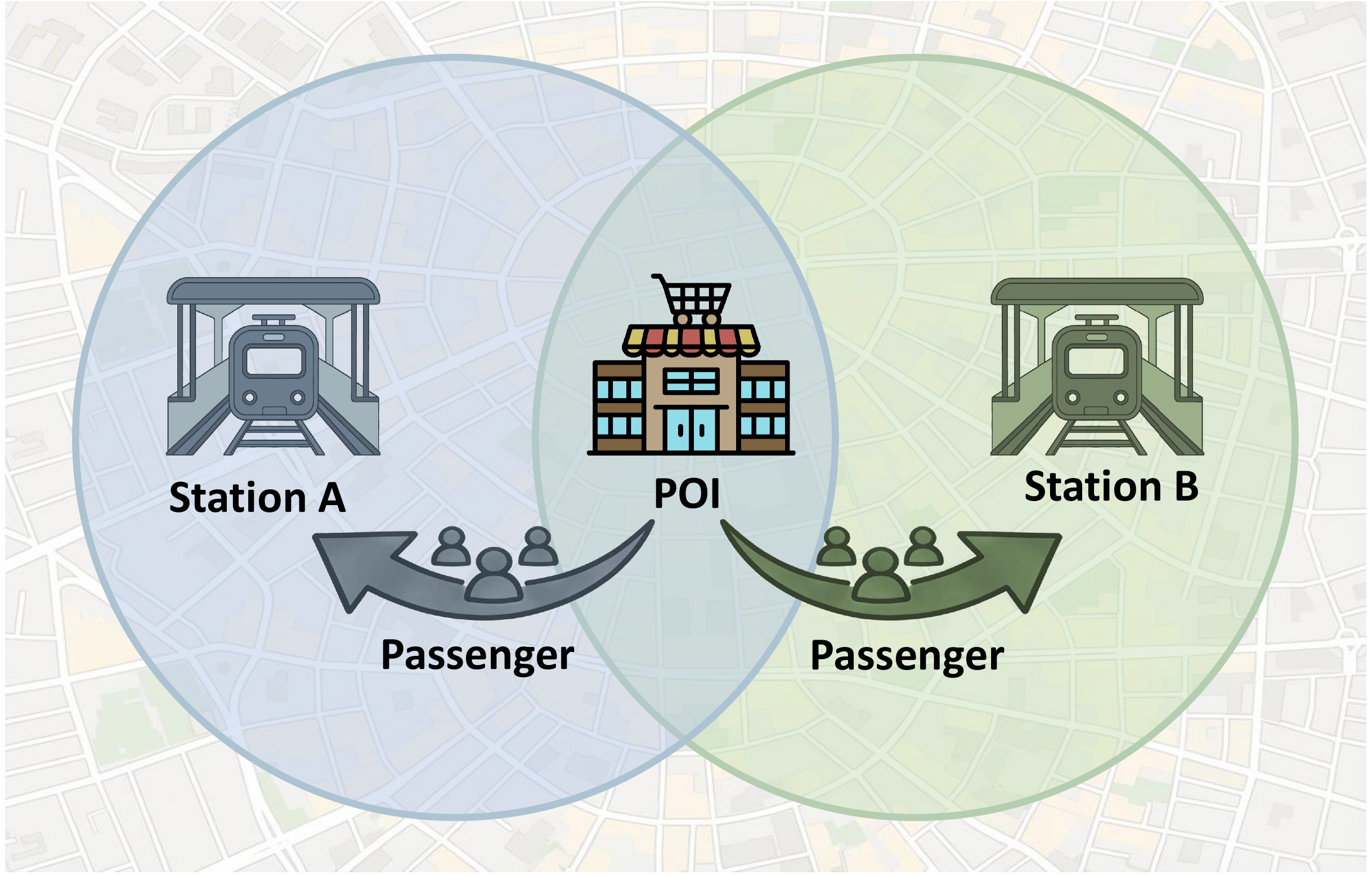}
\vspace{-1mm}
\caption{An example of the interactions among two stations and a POI, implying the ridership of both stations mutually influences each other.}
\label{fig:example}
\vspace{-3mm}
\end{figure}

As key components of rail transit systems, rail transit stations are critical transport hubs which have been widely studied for their impact on urban development~\cite{jiang_understanding_2025}. Traditional research on rail transit stations often relies on manually collected data~\cite{GCYZ202302010} and focuses on a limited number of stations~\cite{CSHJ202201014}, thus fails to capture common features of the stations across the entire rail transit system, and cannot support city-level rail transit station related tasks (e.g., ridership prediction). Consequently, recent studies have shifted toward using large-scale data sources~\cite{app12083867,WANG2025104440} to conduct rail transit station analysis. However, they often treat rail transit stations, points of interest (POIs), and etc. as independent entities, neglecting complex interactions among various urban entities, which limits their application in real-world scenarios. For example, as shown in Figure~\ref{fig:example}, 
station $A$ and station $B$ are adjacent on the same rail transit line, and the walking times from a certain POI to both stations are similar. Passengers departing from the POI may choose either station for boarding. Therefore, the ridership of both stations mutually influences each other, i.e., an increase in the ridership of station $A$ may probably cause a corresponding decrease in the ridership of station $B$. Explicitly modeling such interactions among the POI and both stations could benefit the task of ridership prediction.

Actually, rail transit station related urban entities are distributed across heterogeneous data sources, and it is non-trivial to integrate different kinds of entities~\cite{bampi2025ontology} and model their interactions. Recently, knowledge graphs (KGs) have been adopted for organizing urban data due to their ability to model complex relations among urban entities~\cite{NEURIPS2023_c4a30a4d,10.1145/3746252.3761638}, but current studies do not involve rail transit stations, which prevents their application in city-level rail transit station related tasks.

\begin{figure*}[htbp]
    \centering
    \includegraphics[width=1\textwidth]{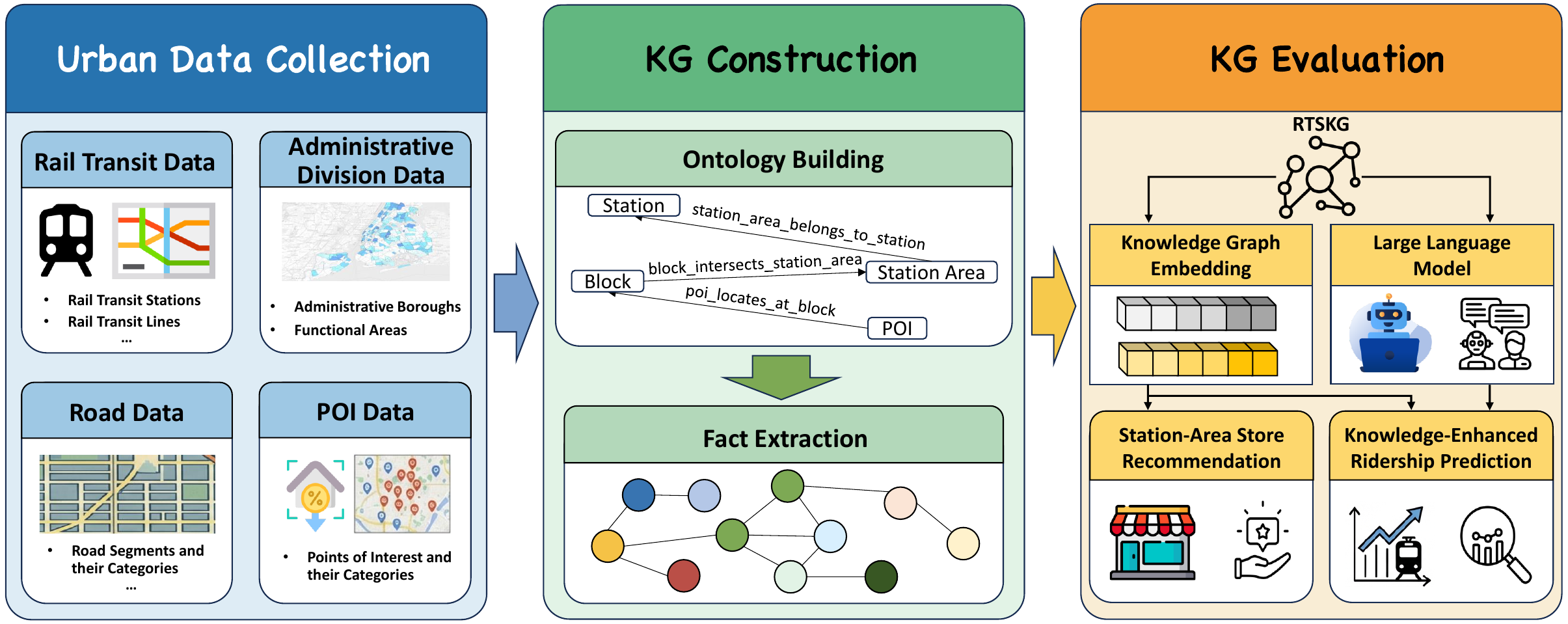} 
    \caption{The overview of our work. Different kinds of urban data are collected to construct RTSKG with a unified schema, and the built KG is evaluated by station-area store recommendation and knowledge-enhanced ridership prediction.
    }
    \label{fig:overview}
\end{figure*}

Thus, in this paper, we propose \textbf{RTSKG}, the \underline{\textbf{R}}ail \underline{\textbf{T}}ransit \underline{\textbf{S}}tation \underline{\textbf{K}}nowledge \underline{\textbf{G}}raph dataset, which explicitly models the spatial and semantic interactions among different kinds of urban entities. 
Specifically, RTSKG consists of two sub-KGs for New York City and Chicago, respectively. For each city, as shown in the overview of our work (Figure~\ref{fig:overview}), we first collect different kinds of urban data, including rail transit data, administrative data, road data, and POI data.
We then build the KG by extracting entities and relations from such collected data with a unified schema which models important classes and relations relevant to rail transit systems. We finally conduct experiments on the designed city-level rail transit station related tasks to show the effectiveness of RTSKG. 

\spara{Contributions.} The main contributions of this paper are summarized as follows:
\begin{itemize}
    \item We propose RTSKG, the first rail transit station knowledge graph dataset, which explicitly models the spatial and semantic interactions among different urban entities with a specially designed ontology as its unified schema.
    \item We design two representative city-level rail transit station related tasks: station-area store recommendation and knowledge-enhanced ridership prediction, which are used to evaluate the practical utility of RTSKG.
    \item We conduct comprehensive experiments on the designed city-level rail transit station related tasks. The results show that leveraging RTSKG for the above tasks has the best performance compared with using existing urban KGs, reflecting the value of RTSKG in city-level rail transit station analysis.
\end{itemize}

\section{Related Work}

\subsection{Rail Transit Station Analysis}
Rail transit station analysis has been a long-standing research topic. Traditional research often relies on manually collected data and covers a limited number of stations. For example, Wu et al. \cite{WU2022103895} utilizes manually measured meteorological data for anthropogenic heat estimating, Nafi et al. \cite{designs5040061} focuses on Hamad Hospital station in Doha and investigates its influence on the surrounding areas. 
Such small-scale data cannot support city-level rail transit station related tasks.
Recent studies try to use large-scale data sources for rail transit station analysis. For example, Yu et al. \cite{YU2022103299} collects large-scale POI data to characterize rail transit stations at the city scale. Lee et al. \cite{app12083867} extracts a vast amount of smart card data from passengers who use rail transit for their trips, and predicts the ridership of rail transit stations based on such data. Zhu et al. \cite{doi:10.1061/JTEPBS.TEENG-7808} constructs the environmental factors of stations based on POI for ridership prediction. However, these data often simply treat urban entities like POIs as independent entities, neglecting complex interactions among heterogeneous urban entities, and we aim to solve this issue in this paper by building a large-scale knowledge graph dataset RTSKG. It explicitly models the spatial and semantic interactions among different kinds of urban entities to facilitate city-level rail transit analysis. 

\subsection{Resources for Urban Modeling}
Urban KGs model relations among urban entities, providing a structured way to integrate and organize such heterogeneous urban data, and have been applied to different urban tasks. For example, among non-public urban KGs, STKG~\cite{10.1145/3494993} jointly models category information of venues and temporal information, and is used for mobility prediction; 
Knowsite~\cite{10.1145/3589132.3625640} constructs a KG which captures cities' key elements and complex relationships for site selection; KG-MUP~\cite{10.1145/3596604} builds a KG with domain entities and applies it for user profile inference. Recently, several open urban KGs have been built. For example, UUKG~\cite{NEURIPS2023_c4a30a4d} is the first open urban KG dataset for knowledge-enhanced urban spatiotemporal predictions,
and HUSK~\cite{10.1145/3746252.3761638} further designs a function zone construction method to capture fine-grained semantic information between POIs and areas, enriching the semantic and spatial relationships in urban KGs. 
However, these urban KGs do not involve rail transit station related information, so they cannot effectively deal with city-level rail transit station related tasks, which can be addressed by our built RTSKG.

Alongside these urban KGs developed for urban tasks, several resources provide representations of knowledge about public transport, geography, and urban infrastructure, thereby guiding the modeling of rail transit systems. For example, the NEPTUNE and Passim datasets~\cite{DBLP:conf/wod/PluS12} are represented in RDF format based on specially developed ontologies and published as Linked Data, describing public transport routes and relevant information on transport services, respectively. LinkedGeoData~\cite{DBLP:journals/semweb/StadlerLHA12} transforms and represents OpenStreetMap data adhering to the RDF data model. GeoKG~\cite{su131910602} uses schema and data layers to represent and integrate geographical knowledge from multi-source data.
City Infrastructure Ontologies~\cite{DBLP:journals/urban/DuWDMCCETCSRC23} represent city infrastructure assets and their interdependencies in OWL 2. LibCity-Dataset~\cite{10.1093/iti/liad021} standardizes urban spatial-temporal data storage using atomic files that represent geographic entities and relations.
However, these resources either focus on public transport knowledge with limited coverage of other urban entities relevant to rail transit stations, or represent urban knowledge in a generic manner without specifically organizing entities and relations around rail transit systems. RTSKG addresses this gap by integrating different kinds of urban entities under a specially designed schema.

\section{Knowledge Graph Construction}

In this section, we introduce different kinds of urban data we collect. We also present our designed ontology and explain how we extract facts from the collected data to build the KG.

\subsection{Urban Data Collection}
\label{data}
We collect the urban data grouped into four categories for two metropolises: New York City and Chicago.

\dataparagraph{Rail Transit Data.} 
The rail transit data involve the information of key components of rail transit systems and it contains the data of rail transit stations, rail transit lines, and rail transit station entrances \& exits. Rail transit stations and rail transit lines are obtained from the Open NY\footnote{\url{https://data.ny.gov/}} and CHI Data Portal\footnote{\url{https://data.cityofchicago.org/}}. We collect station entrances \& exits from Open NY and OpenStreetMap\footnote{\url{https://www.openstreetmap.org/}} (OSM) for New York City and Chicago, respectively.

\dataparagraph{Administrative Division Data.}
Administrative divisions represent hierarchical geographic regions created by governments. We collect two types of administrative divisions, i.e., administrative boroughs and functional areas. Administrative boroughs represent large-scale administrative divisions that support city management and public service delivery. Functional areas refer to the subdivisions of a city according to their functions, such as residential areas. 
The administrative division data are obtained from NYC Opendata\footnote{\url{https://opendata.cityofnewyork.us/}} and CHI Data Portal. 

\dataparagraph{Road Data.} 
Road data involves information about the urban road network including road segments, road categories and roadbeds. We collect road segments and road categories for New York City and Chicago from NYC OpenData and the Chicago Data Portal, respectively. Roadbeds are the regions representing the width of road segments, which can be applied to generate urban entities like urban blocks~\cite{Ebrahimi20} for analysis. We collect roadbeds directly from NYC OpenData for New York City and generate roadbeds based on self-defined widths of road segments for Chicago.

\dataparagraph{POI Data.} 
POIs are specific places where people gather and conduct daily activities~\cite{LIU2020102610} and they are widely used to support city-level urban tasks. The POI data are obtained from the famous urban KG dataset UUKG~\cite{NEURIPS2023_c4a30a4d}.

\subsection{Ontology Building \& Fact Extraction}
This subsection introduces the construction process of KG.

\subsubsection{Ontology Building}
\begin{figure}[t]
\centering
  \includegraphics[width=0.9\linewidth]{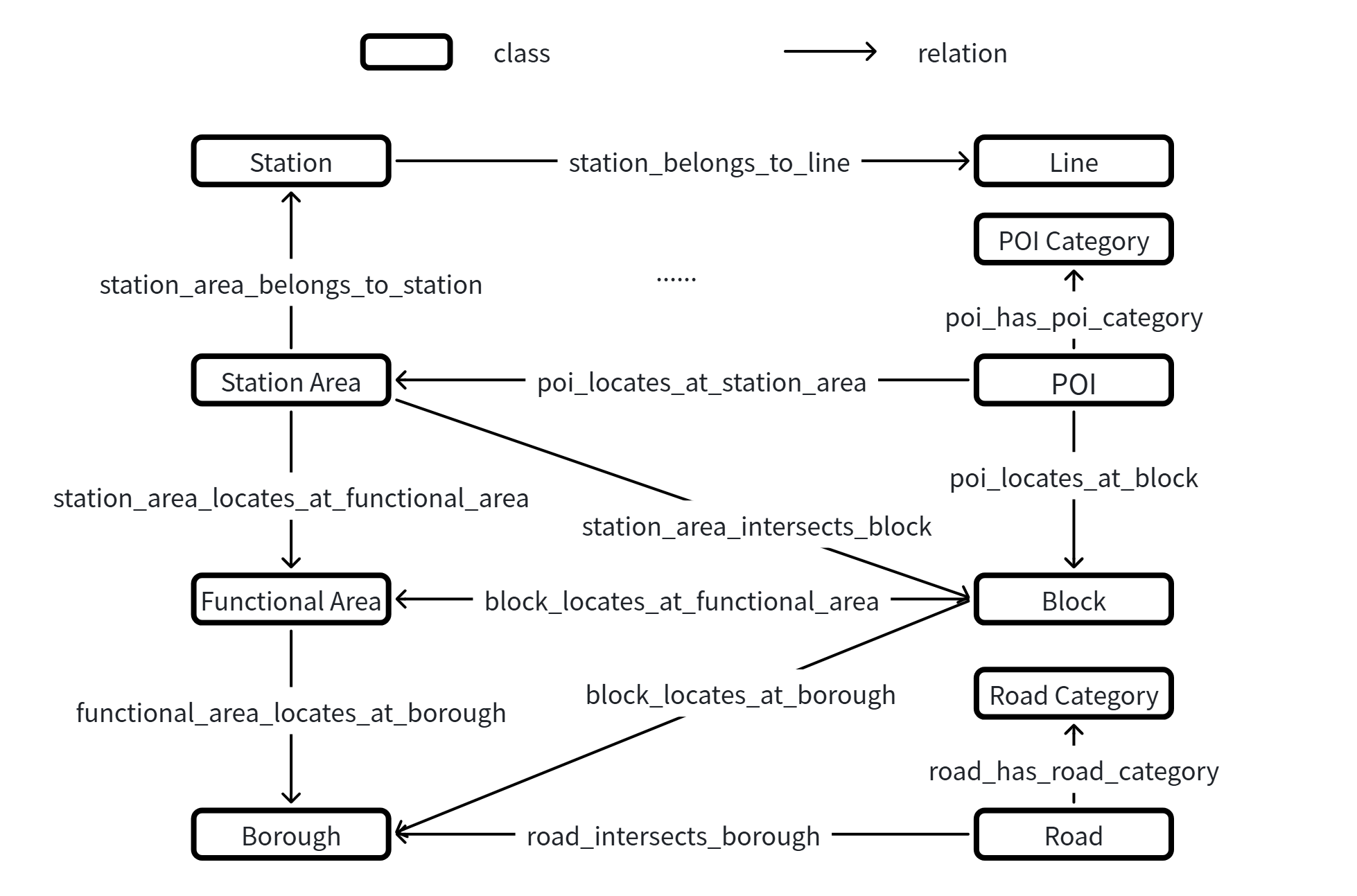}
  \caption{A part of the RTSKG ontology.}
  \label{fig:sch}
\end{figure}

We build the RTSKG ontology as the schema~\cite{:/DI/doi/10.3724/2096-7004.di.2025.0001} to guide extracting and integrating factual knowledge from
different kinds of urban data.
Following the broadly accepted procedure~\cite{reason:NoyMcG01a,10.1007/978-3-031-47243-5_4}, we first determine the domain and scope of the ontology. RTSKG is designed to model the interactions among different kinds of urban entities and support city-level rail transit station related tasks. Based on concrete needs in urban analysis, we formulate a set of competency questions (listed in our GitHub repository) to specify the important concepts to be covered by the ontology.
We then consider reusing existing vocabularies. For example, we reuse the standard RDF, RDFS, OWL and Dublin Core vocabularies, including \path{rdf:type} linking from instances to classes, \path{rdfs:label} recording the labels of classes, relations (i.e., object properties) and attributes (i.e., data properties), \path{owl:versionIRI} identifying the IRI used to identify the version of the ontology, \path{dcterms:description} providing a textual description of the ontology, \path{rdfs:domain} specifying that any resource that has a given relation or attribute is an instance of one or more classes, and \path{rdfs:range} specifying that the values of a relation or attribute are instances of one or more classes. We then define the classes, relations, and attributes in the RTSKG ontology, where attributes are used to describe basic spatiotemporal information of urban entities (e.g., coordinates of POIs). Due to space limitations, this subsection focuses on the classes and relations of RTSKG, while the full ontology is available online: \url{https://w3id.org/rtskg/ontology}. Figure~\ref{fig:sch} shows a part of our specially designed ontology.

We define ten classes as follows. 
(1) \textit{Borough} refers to administrative borough describing high-level administrative boundaries of a city; 
(2) \textit{Functional Area} represents the subdivision of a city.  Along with \textit{Borough}, it enables urban entities from different data sources to be aligned according to administrative divisions;
(3) \textit{Station} refers to the rail transit station, which serves as a transport hub for passengers to board or alight from the rail transit vehicles like subway train;
(4) \textit{Station Area} refers to the region accessible from a rail transit station within a specified walking time, providing a local area around the station for analysis;  
(5) \textit{Line} refers to the rail transit line, which is represented as an abstract sequence of connected stations along the route;
(6) \textit{Road} denotes the road segment in a city, which forms the traffic network and provides a basis for human mobility, potentially supporting city-level rail transit station related tasks;
(7) \textit{Block} refers to the urban block, which is a finer-grained
spatial unit that constitutes the fundamental element of the physical
structure of urban areas and supports urban analysis;
(8) \textit{POI} represents the basic functional units in a city, which are widely used to support city-level urban tasks;
(9) \textit{POI Category} refers to category of POI, which characterizes the function of POI; 
(10) \textit{Road Category} is the category determined by the hierarchy or function of the road.

We also define nineteen relations, which are further categorized into four types as follows.

\paragraph{Geographic Location.} 
    This relation type models that one instance locates at another instance. We define seven geographic location relations in RTSKG. Specifically,
    \path{poi_locates_at_station_area} and \path{poi_locates_at_block} represent location relations between point and polygonal instances. \path{station_area_locates_at_borough}, \path{station_area_locates_at_functional_area}, \path{block_locates_at_borough}, \path{block_locates_at_functional_area} and \path{functional_area_locates_at_borough} represent location relations between two polygonal instances.
    \paragraph{Geographic Adjacency.} This relation type models that one instance is adjacent to another instance, and we define two geographic adjacency relations in RTSKG: \path{borough_is_adjacent_to_borough} and \path{functional_area_is_adjacent_to_functional_area}.  Both relations enable the ontology to model the adjacency between two polygonal instances.
    \paragraph{Geographic Intersection.}  This relation type models that one instance intersects another instance, and we define six geographic intersection relations in RTSKG. Specifically, \path{road_intersects_borough}, \path{road_intersects_functional_area}, and \path{road_intersects_station_area} model the intersections between linear and polygonal instances, characterizing the spatial distribution of the road network across geographic regions. \path{station_area_intersects_block}, \path{station_area_intersects_station_area}, and \path{block_intersects_station_area} represent intersections between two polygonal instances, capturing potential interactions through their shared spatial regions (e.g., the region of overlap between two station areas).
    \paragraph{Non-Geographic Relation.} 
    This relation type covers the relations other than the above three relation types. We define four non-geographic relations as follows: \path{station_area_belongs_to_station}, \path{station_belongs_to_line},  \path{poi_has_poi_category}, and \path{road_has_road_category}. These relations do not capture spatial interactions between instances, but are still important in rail transit station analysis.

\subsubsection{Fact Extraction}
We extract facts from the data collected in Section~\ref{data}. To handle heterogeneous urban data, we employ the GeoPandas~\cite{kelsey_jordahl_2020_3946761}, a Python library for dealing with geospatial data, to extract urban entities and facts.
Specifically, we use GeoPandas to extract the instances belonging to the classes \textit{Borough}, \textit{Functional Area}, \textit{Station}, \textit{Road}, and \textit{POI}, together with their spatial locations, from the geospatial data in the shapefile format. We also obtain the instances belonging to the classes \textit{Line}, \textit{POI Category}, and \textit{Road Category} directly from the tabular data we collect with the comma-separated values format.

\begin{figure}[t]
\centering
  \includegraphics[width=0.76\linewidth]{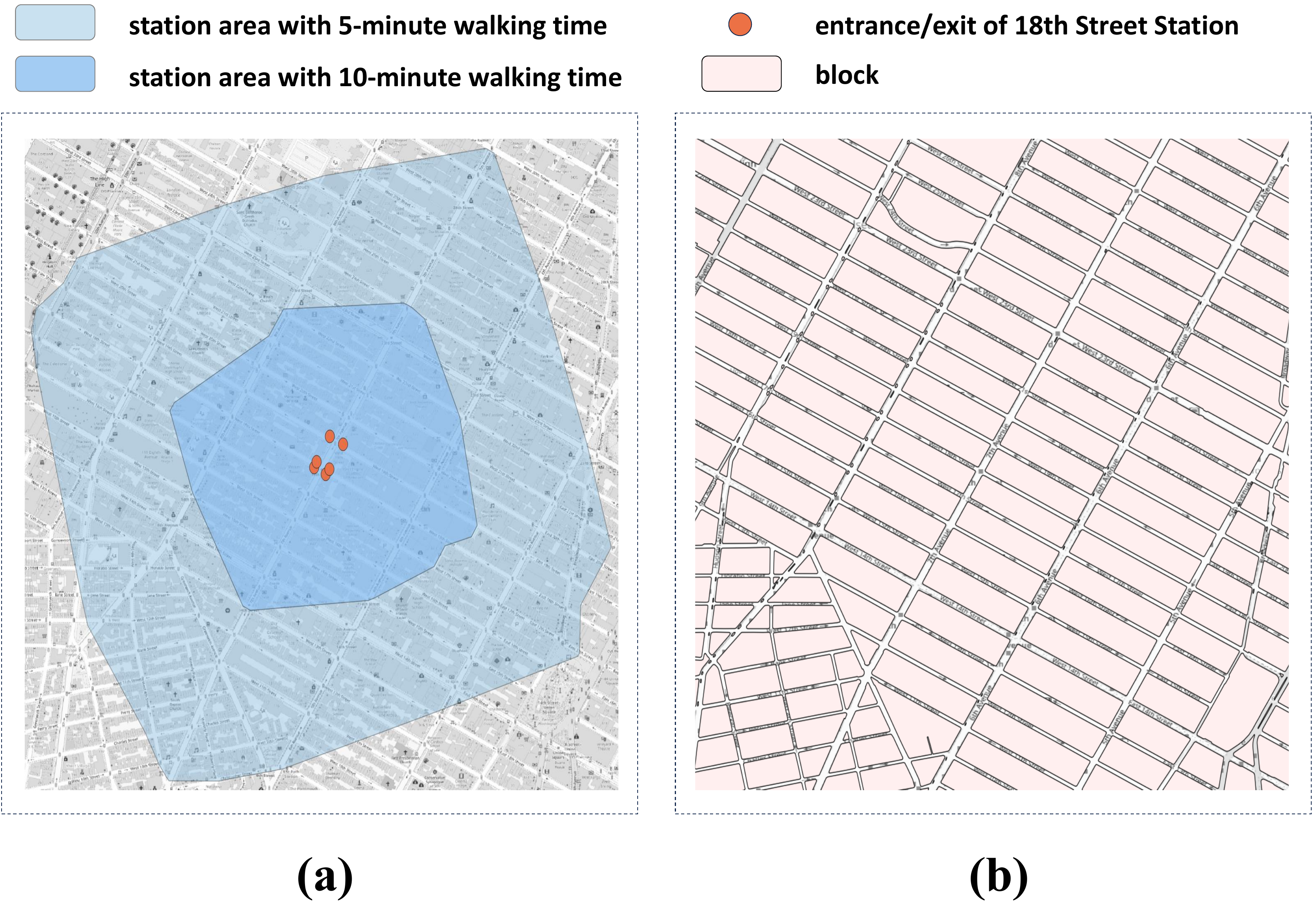}
  \caption{An example of station areas and blocks in New York City. (a) Station areas of 18th Street station, where the inner polygon represents the station area with 5-minute walking time, the outer polygon represents the station area with 10-minute walking time, and the nodes represent the entrances or exits of 18th Street Station. (b) A part of Blocks.}
  \label{fig:isoblock}
\end{figure}

For the instances in the class \textit{Station Area}, following the definition in~\cite{TYNDALL2022103411}, we construct a station area for each station based on the isochrone regions (i.e., the region can be reached within a certain walking time) of its entrances or exits. For each entrance or exit, we generate an isochrone region using the Mapbox Isochrone API\footnote{\url{https://www.mapbox.com/}} based on its geographic coordinates. The station area of a station is then obtained by taking the union of the isochrone regions of all the entrances and exits. In this paper, we construct station areas with two fixed walking times, i.e., 5 minutes and 10 minutes, which are common thresholds in transit-oriented development (TOD) and station-area studies~\cite{Yajie2024,Curtis01082008}, for each station. Figure~\ref{fig:isoblock} (a) shows the station areas of the 18th Street station with 5-minute and 10-minute walking times. For the instances in the class \textit{Block}, we define them as the regions in the city excluding roadbeds. Figure~\ref{fig:isoblock} (b) visualizes a part of blocks in New York City. After obtaining the instances of all classes, we use GeoPandas to extract all relations of geographic location, geographic adjacency, and geographic intersection. For non-geographic relations, we directly obtain them from the collected data with the comma-separated values format.

By integrating the extracted facts, we build KGs for New York City and Chicago, respectively. Table~\ref{dataset} shows the statistics of the KGs for the cities over the classes, relations, and attributes defined in the RTSKG ontology, while the dumps released on Zenodo additionally provide triples with predicates from standard vocabularies. The KG construction process can be extended to other cities by collecting the required urban data and adapting the Python scripts to handle differences in data formats.
All class URIs
in the namespace \url{https://w3id.org/rtskg/ontology/class/} 
and all instance URIs in the namespaces \url{https://w3id.org/rtskg/new_york_city/instance/} and \url{https://w3id.org/rtskg/chicago/instance/} are dereferenceable.

\begin{table}[htbp]
    \centering
        \caption{The statistics of RTSKG.}
    \begin{tabular}{c| c c c c}
        \toprule
        City & \#Instances & \#Relations & \#Relation Triples & \#Attribute Triples \\
        \midrule
        New York City & 238,839 & 19 & 1,113,638 & 239,090 \\
        Chicago & 151,845 & 19 & 448,716 & 151,943\\
        \bottomrule
    \end{tabular}
        \label{dataset}
\end{table}

\section{Knowledge Graph Embedding}
In this section, we apply knowledge graph embedding (KGE) models to RTSKG and select the best KGE model on link prediction for city-level rail transit station related tasks (details will be given in Section \ref{Recommendation} and Section \ref{ridership_task}).

\paragraph{Problem Formulation.} 
Given a KG $\mathcal{G}$, KGE is to learn vector representations (i.e., embeddings) for instances and relations in a dense and low-dimensional space. The embedding of the instance $i\in\mathcal{I}$ is denoted as $\bm{i}$ and the embedding of the relation $r\in\mathcal{R}$ is denoted as $\bm{r}$, where $\mathcal{I}$ and $\mathcal{R}$ are respectively the sets of instances and relations.

\subsection{Experimental Setup}
\label{gie_setup}
We compared the performance of different KGE models on link prediction which aims to predict the missing tail or head instance for a query $(h, r, ?)$ or $(?, r, t)$, where $h,t \in \mathcal{I}$ denote the head and tail instances, respectively.
\paragraph{Evaluation Protocol.}  We reported the following evaluation metrics for link prediction: (1) Hits@$k$ ($k$=1,3), the percentage of correct instances in the top-$k$ ranked instances. (2) Mean reciprocal rank (MRR), the average multiplicative inverse of the ranks for all instances.

\paragraph{Models.} We employed various KGE models to represent KGs, and chose the best model as the base model for city-level rail transit station related tasks. The models are listed as follows:
(1) TransE~\cite{NIPS2013_1cecc7a7} is the classic translation-based KGE model. (2) DistMult~\cite{yang2015embedding} is a  bilinear KGE model with tensor decomposition. (3) ComplEx~\cite{pmlr-v48-trouillon16} is an extension of DistMult which learns KG embeddings in a complex space. (4) RotatE~\cite{DBLP:conf/iclr/SunDNT19} is a KGE model which defines each relation as a rotation in the complex vector space. (5) MuRE~\cite{DBLP:conf/nips/BalazevicAH19} is a translational distance KGE model with a diagonal relational matrix. (6) TuckER~\cite{DBLP:conf/emnlp/BalazevicAH19} is a KGE model based on tucker decomposition of the binary tensor representation of triples. (7) QuatE~\cite{DBLP:conf/nips/0007TYL19} is a KGE model which introduces more expressive hypercomplex representations to model entities and relations. 
(8) AttH~\cite{DBLP:conf/acl/ChamiWJSRR20} is a KGE model which combines hyperbolic reflections and rotations with attention to learn embeddings in KG.
(9) RefH/RotH~\cite{DBLP:conf/acl/ChamiWJSRR20} are variants of AttH, and they are  hyperbolic KGE models with reflection and rotation, respectively. (10) GIE~\cite{DBLP:conf/aaai/CaoX0CH22} is a product space KGE model, which aims to learn spatial structures interactively between the Euclidean, hyperbolic and hyperspherical spaces.

\paragraph{Implementation Details.}
All models were implemented by PyTorch, and all experiments were conducted on an RTX 3090 GPU. Following the previous work~\cite{NEURIPS2023_c4a30a4d}, we fixed the negative sampling size as 50 and the embedding dimension as 32 for all models.

\subsection{Results}
\label{GIE}
Table~\ref{embedding} reports the link prediction results in the sub-KGs for New York City and Chicago. Experimental results indicate that GIE outperforms other KGE models across all link prediction metrics on both sub-KGs. This is mainly because our ontology contains both hierarchical structures and cyclic structures. As shown in Figure~\ref{fig:sch}, the relations among \textit{Station Area}, \textit{Functional Area}, and \textit{Borough} form a hierarchical structure, and the relations among \textit{Station Area}, \textit{POI}, and \textit{Block} form a cyclic structure. GIE is designed to capture such structures simultaneously~\cite{NEURIPS2023_c4a30a4d}. Motivated by GIE’s superior KGE ability, we chose GIE as the base model to generate instance and relation embeddings for city-level rail transit station related tasks.

\begin{table}[h!]
\centering
\small
\setlength{\tabcolsep}{3.5pt}

\caption{Link prediction results of eleven KGE models on the sub-KGs for New York City and Chicago.}

\begin{tabular}{c|ccc|ccc}
\midrule 
\multirow{2}{*}{Model} & \multicolumn{3}{c|}{New York City} & \multicolumn{3}{c}{Chicago} \\
 & Hits@1 & Hits@3 & MRR & Hits@1 & Hits@3 & MRR \\ 
\midrule
TransE  & 0.291 & 0.417 & 0.378 & 0.272 & 0.394 & 0.355\\
DistMult & 0.302 & 0.450 & 0.389 & 0.196 & 0.349 & 0.289 \\
MuRE & 0.325 & 0.460 & 0.413 & 0.308 & 0.430 & 0.388\\
TuckER  & 0.200 & 0.351 & 0.297 & 0.185 & 0.316 & 0.272 \\
RotatE & 0.140 & 0.304 & 0.250 & 0.079 & 0.190 & 0.169\\
ComplEx & 0.208 & 0.384 & 0.315 & 0.154 & 0.305 & 0.250\\
QuatE & 0.291 & 0.443 & 0.381 & 0.237 & 0.377 & 0.322 \\
RotH & 0.304 & 0.439 & 0.395 & 0.277 & 0.404 & 0.365\\
RefH & 0.304 & 0.435 & 0.394 & 0.277 & 0.401 & 0.363 \\
ATTH & 0.319 & 0.463 & 0.412 & 0.310 & 0.432 & 0.392\\
\midrule
\textbf{GIE} & \textbf{0.326} & \textbf{0.467} & \textbf{0.417} 
& \textbf{0.317} & \textbf{0.438} & \textbf{0.397}\\
\midrule
\end{tabular}
\label{embedding}

\end{table}

\section{Station-Area Store Recommendation}
\label{Recommendation}
In this section, we introduce the task of station-area store recommendation, which aims to evaluate the utility of our RTSKG for the analysis of economic development in the surrounding areas of stations.
\paragraph{Problem Formulation.} Station-area store recommendation can be formally modeled as a special link prediction task.
Given a \textit{POI} instance $h_p$ located in a station area, station-area store recommendation aims to predict the category through the query $(h_p,r_p,?)$, where $r_p$ denotes the relation \path{poi_has_poi_category}.

\subsection{Experimental Setup}
We conducted station-area store recommendation in New York City and Chicago.
We still selected MRR and Hits@$k$ ($k$=1,3) as the evaluation metrics. We compared RTSKG with two existing open urban KG datasets, i.e.,  UUKG~\cite{NEURIPS2023_c4a30a4d} and HUSK~\cite{10.1145/3746252.3761638}.

\paragraph{Data Description.} For each city, we sampled POIs located at station areas with 5-minute walking time, and ensured that each sampled POI appears as an instance in each of the three KG datasets (i.e., RTSKG, UUKG, and HUSK). We then constructed the validation and test sets by extracting relation triples with the relation \path{poi_has_poi_category} whose head instances are these sampled POIs. All remaining relation triples were used as the training set for each KG dataset.

\paragraph{Implementation Details.} 
We randomly sampled 2,500 relation triples and split them into 500 relation triples for validation and 2,000 relation triples for test. We employed GIE as the base model, and we adopted the same hyper-parameter settings as in Section~\ref{gie_setup}.

\subsection{Results} 
As shown in Table~\ref{store}, GIE achieves the best Hits@$k$ and MRR when leveraging RTSKG, compared with leveraging other KGs. By performing the task of station-area store recommendation with RTSKG, we can more accurately identify the most suitable development directions for stores in station areas, which enables more efficient resource allocation and contributes to the economic development of the surrounding areas of the stations.

\begin{table}[t]
\centering
\small
\setlength{\tabcolsep}{3.5pt}

\caption{The results on station-area store recommendation.}

\begin{tabular}{c|ccc|ccc}

\midrule
\multirow{2}{*}{KG} & \multicolumn{3}{c|}{New York City} & \multicolumn{3}{c}{Chicago} \\
 & Hits@1 & Hits@3 & MRR & Hits@1 & Hits@3 & MRR \\ 
\midrule
UUKG  & 0.698 & 0.775 & 0.757 & 0.264 & 0.533 & 0.455\\
HUSK  & 0.705 & 0.785 & 0.770 & 0.356 & 0.615 & 0.524\\
\midrule
\textbf{RTSKG}  
& \textbf{0.720} & \textbf{0.794} & \textbf{0.782}
& \textbf{0.380} & \textbf{0.694} & \textbf{0.570} \\
\midrule
\end{tabular}
\label{store}
\end{table}

\section{Knowledge-Enhanced Ridership Prediction}
\label{ridership_task}
In this section, we introduce the task of knowledge-enhanced ridership prediction, which aims to evaluate the practical utility of our RTSKG for the analysis of rail transit station related urban mobility.
\paragraph{Problem Formulation.} Ridership prediction is a classic urban task, and it is typically formulated as time series prediction, which aims to predict the station ridership on the next day based on the past station ridership sequence.
Knowledge-enhanced ridership prediction aims to perform ridership prediction by incorporating the KG for knowledge enhancement.
Let $X^{(t)} \in \mathbb{R}^{a}$ denote the ridership observations at the time step $t$, where $a$ denotes the number of rail transit stations. Given the past observations $(X^{(t-T)}, X^{(t-T+1)},  \ldots, X^{(t-1)})$, and an external KG $\mathcal{G}$, where $T$ is the input length of the past observations, knowledge-enhanced ridership prediction aims to employ a model $m$ to predict $X^{(t)} = m\left( (X^{(t-T)}, X^{(t-T+1)}, \ldots, X^{(t-1)}), \mathcal{G} \right)$, where $m(\cdot)$ is the function that model $m$ transforms a sequence of ridership to a single ridership value.
Depending on the type of model adopted, knowledge-enhanced ridership prediction can be further categorized into traditional knowledge-enhanced ridership prediction and LLM-based knowledge-enhanced ridership prediction.

\paragraph{Traditional Knowledge-Enhanced Ridership Prediction.} 
Traditional knowledge-enhanced ridership prediction~\cite{NEURIPS2023_c4a30a4d,10.1145/3746252.3761638} typically performs knowledge enhancement by using KG embeddings as external knowledge features and directly concatenating them with past ridership observations as the input of the model $m$.

\paragraph{LLM-based Knowledge-Enhanced Ridership Prediction.} 
Recently, large language models (LLMs) have shown impressive performance on various  tasks, and applications combining LLMs with KGs have become a significant research direction as well. Motivated by this, we followed the previous work~\cite{10356715} and proposed the task of LLM-based knowledge-enhanced ridership prediction. Specifically, we verbalized each time series instance (i.e., the ridership observations over the past days of a station) into a natural language query, e.g., “From September 05 to September 16, 2025, the rail transit station `Myrtle-Wyckoff Avs' located at the `Bushwick North' functional area recorded the following ridership numbers: 18030 12496 ... 18461 on each day. What is the ridership going to be on September 17, 2025?”. Meanwhile, we verbalized each KG relation triple into a natural language statement, e.g., ``The rail transit station `Myrtle-Wyckoff Avs' belongs to the rail transit line `Canarsie'''. We then encoded all the queries and verbalized relation triples as embeddings with a pretrained language model, and computed similarity scores between the queries and relation triples by cosine similarities of their embeddings. For each query, we retrieved the top-$n$ most relevant verbalized relation triples from the KG based on these similarity scores. Next, we used an LLM to summarize the retrieved relation triples into an abstract, which serves as external knowledge for the query. Finally, we concatenated the query with the abstract to form a prompt and fed it into another LLM to perform knowledge-enhanced ridership prediction. The prompts are provided in our GitHub repository.
\subsection{Experimental Setup}
We conducted knowledge-enhanced ridership prediction in New York City and Chicago.
\paragraph{Evaluation Protocol.} For knowledge-enhanced ridership prediction, we chose evaluation metrics as follows: Mean Absolute Error (MAE): the mean absolute error of predicted ridership and ground ridership, and Root Mean Squared Error (RMSE): the root mean squared error of predicted ridership and ground ridership.

\paragraph{Data Description.} We collected daily ridership records of rail transit stations from U.S. Government's Open Data\footnote{\url{https://data.gov/}}, and we manually filtered out the records that may be ambiguous or erroneous. 
The statistics of filtered ridership data are summarized in Table \ref{ridership_data}.

\paragraph{Models.} For traditional knowledge-enhanced ridership prediction, we employed five representative spatiotemporal prediction models: (1) STGCN~\cite{DBLP:conf/ijcai/YuYZ18} is spatiotemporal graph convolutional network, which has complete convolutional structures. (2) ASTGCN~\cite{DBLP:conf/aaai/GuoLFSW19} is attention based spatial-temporal graph convolutional network, which combines the spatial-temporal attention mechanism and the spatial-temporal convolution. 
(3) TGCN~\cite{DBLP:journals/tits/ZhaoSZLWLDL20} is a graph convolutional network, which is combined with the gated recurrent unit. (4) MSTGCN~\cite{DBLP:conf/aaai/GuoLFSW19} is multi-component spatial-temporal graph convolution network, which is a variant of ASTGCN.  (5) STPGCN~\cite{9945663} is spatiotemporal position-aware graph convolutional network, which constructs a trainable embedding module and represents the spatial and temporal positions of the nodes in graph. For LLM-based knowledge-enhanced ridership prediction, we adopted GPT-4o\footnote{\url{https://openai.com/index/hello-gpt-4o/}}, a widely adopted LLM provided by OpenAI, as the base LLM.

\paragraph{Baselines.} 
We compared the models enhanced by RTSKG with the variants without knowledge enhancement (denoted as Vanilla). We also evaluated two existing urban KGs, UUKG and HUSK, by leveraging them for knowledge enhancement. For traditional knowledge-enhanced ridership prediction, since neither existing KG contains rail transit stations, we applied their instances from commonly adopted class (i.e., \textit{Functional Area}) for knowledge enhancement, denoted as w/UUKG and w/HUSK. For fair comparison, we additionally enhanced the model with instances in the class \textit{Station} and the class \textit{Functional Area} from RTSKG, denoted as w/RTSKG-S and w/RTSKG-F, respectively. For LLM-based knowledge-enhanced ridership prediction, we directly used all the KGs as external knowledge bases for knowledge enhancement. The variant without knowledge enhancement is denoted as Vanilla-LLM, and the variants enhanced with RTSKG, HUSK, and UUKG are denoted as w/RTSKG-LLM, w/HUSK-LLM, and w/UUKG-LLM, respectively.

\paragraph{Implementation Details.} In this task, we used the observations from the past 12 days to predict ridership for the next day. For traditional knowledge-enhanced ridership prediction, we split the dataset into training, validation, and test sets with a ratio of 7:1:2. 
All models were trained using Adam~\cite{kingma2015adam} with a batch size of 64. 
For LLM-based knowledge-enhanced ridership prediction, we randomly sampled 500 records from the test set for evaluation. We set the number of retrieved relation triples $n$ as 50 and
we adopted MiniLM~\cite{NEURIPS2020_3f5ee243} to encode queries and verbalized relation triples as embeddings.

\begin{table}[t]
\centering
\small
\setlength{\tabcolsep}{15pt}
\caption{The statistics of ridership data}

\begin{tabular}{c|c|c}
\midrule
City & {New York City} & {Chicago} \\
\midrule
\# of records  & 343368  & 114456 \\ 
\midrule
Time span  & \multicolumn{2}{c}{01/01/2023-09/30/2025}\\
\midrule
\end{tabular}

\label{ridership_data}
\end{table}

\begin{table}[h!]
\centering

\footnotesize
\setlength{\tabcolsep}{3pt}
\renewcommand{\arraystretch}{0.95}

\caption{The comparison results of traditional knowledge-enhanced ridership prediction. The best results are indicated by bold numbers.}

\begin{adjustbox}{max width=\textwidth}
\begin{tabular}{c c cc cc}
\toprule
\multirow{2}{*}{Model} & \multirow{2}{*}{Metric}
& \multicolumn{2}{c}{New York City}
& \multicolumn{2}{c}{Chicago} \\
\cmidrule(lr){3-4} \cmidrule(lr){5-6}
 &  & MAE & RMSE & MAE & RMSE \\
\midrule

\multirow{5}{*}{STGCN}
& Vanilla    & 1490.229 & 2555.250  & 272.275 & 589.112 \\
& w/UUKG     & 1196.467 & 2301.635 & 259.893 & 546.069 \\
& w/HUSK     & 1193.374 & 2322.347 & 250.546 & 544.304\\
& w/RTSKG-F  & 936.359 &1819.504 & 247.839 & 543.506 \\
& w/RTSKG-S  & \textbf{735.471} & \textbf{1563.717} & \textbf{241.169} & \textbf{538.621} \\
\midrule

\multirow{5}{*}{ASTGCN}
& Vanilla    & 101.066 & 218.336    & 61.038 & 160.673 \\
& w/UUKG     & 94.583 & 209.423    & 47.208 & 136.242 \\
& w/HUSK     & 100.330 & 213.399    & 48.003 & 137.948 \\
& w/RTSKG-F  & 91.003 & 181.706    & 44.958 & 131.662 \\
& w/RTSKG-S  &  \textbf{70.766} & \textbf{173.516}    & \textbf{42.143} & \textbf{126.379} \\
\midrule

\multirow{5}{*}{MSTGCN}
& Vanilla    & 208.735 & 404.118 & 87.532 & 185.724 \\
& w/UUKG     & 217.124 & 389.656 & 85.472 & 179.737 \\
& w/HUSK     & 206.557 & 374.751 & 83.794 & 182.502 \\
& w/RTSKG-F  & 193.388 & 366.905 & 81.437 & 179.197 \\
& w/RTSKG-S  & \textbf{176.531} & \textbf{353.559} & \textbf{81.330} & \textbf{178.617} \\
\midrule

\multirow{5}{*}{TGCN}
& Vanilla    & 3724.667 & 6686.524 & 404.930 & 719.092 \\
& w/UUKG     & 3680.524 & 6069.055 & 399.700 & 717.116 \\
& w/HUSK     & 3619.817 & 6242.119 & 393.618 & 698.174 \\
& w/RTSKG-F  & 3383.251 & 6004.507 &  381.304 & 693.866 \\
& w/RTSKG-S  & \textbf{3222.358} & \textbf{5580.707} & \textbf{378.903} & \textbf{687.910} \\
\midrule

\multirow{5}{*}{STPGCN}
& Vanilla    & 1100.163 & 2304.365 & 319.826 & 695.963 \\
& w/UUKG     & 1081.648 & 2170.809  & 319.722 & 681.038 \\
& w/HUSK     & 1088.856 & 2238.719 & 315.819 & 680.975 \\
& w/RTSKG-F  & 1050.264 & 2120.993 & 312.528 & 678.507 \\
& w/RTSKG-S  & \textbf{1007.729} & \textbf{2075.475} & \textbf{307.396} & \textbf{673.780} \\

\bottomrule
\end{tabular}
\end{adjustbox}

\label{ridership_result}

\end{table}

\subsection{Results}
\paragraph{Traditional Knowledge-Enhanced Ridership Prediction.} 
As shown in Table \ref{ridership_result}, the experimental results indicate that knowledge enhancement with instances in either the \textit{Station} class or the \textit{Functional Area} class from RTSKG consistently improves the performance of models on ridership prediction compared to the models without enhancement. Furthermore, the models enhanced with the instances in the class \textit{Station} outperform those enhanced with the instances in the class \textit{Functional Area}, suggesting that explicitly modeling stations in the KG can capture the mutual influence between the ridership of different stations, which is more beneficial for ridership prediction. Additionally, the model enhanced with the instances in the class \textit{Functional Area} from RTSKG achieves better performance compared with those enhanced by the instances in the class \textit{Functional Area} from other KGs, as RTSKG models the interactions between different kinds of urban entities, allowing the embeddings of instances in the class \textit{Functional Area} to implicitly include features from stations. The experimental results fully demonstrate the effectiveness of RTSKG, highlighting that explicitly modeling rail transit stations in the KG helps capture the implicit interactions among urban entities, and helps government departments adjust operations in advance of ridership changes, thus can benefit the analysis of urban mobility. 

\begin{figure}[t]
\centering
  \includegraphics[width=0.85\linewidth]{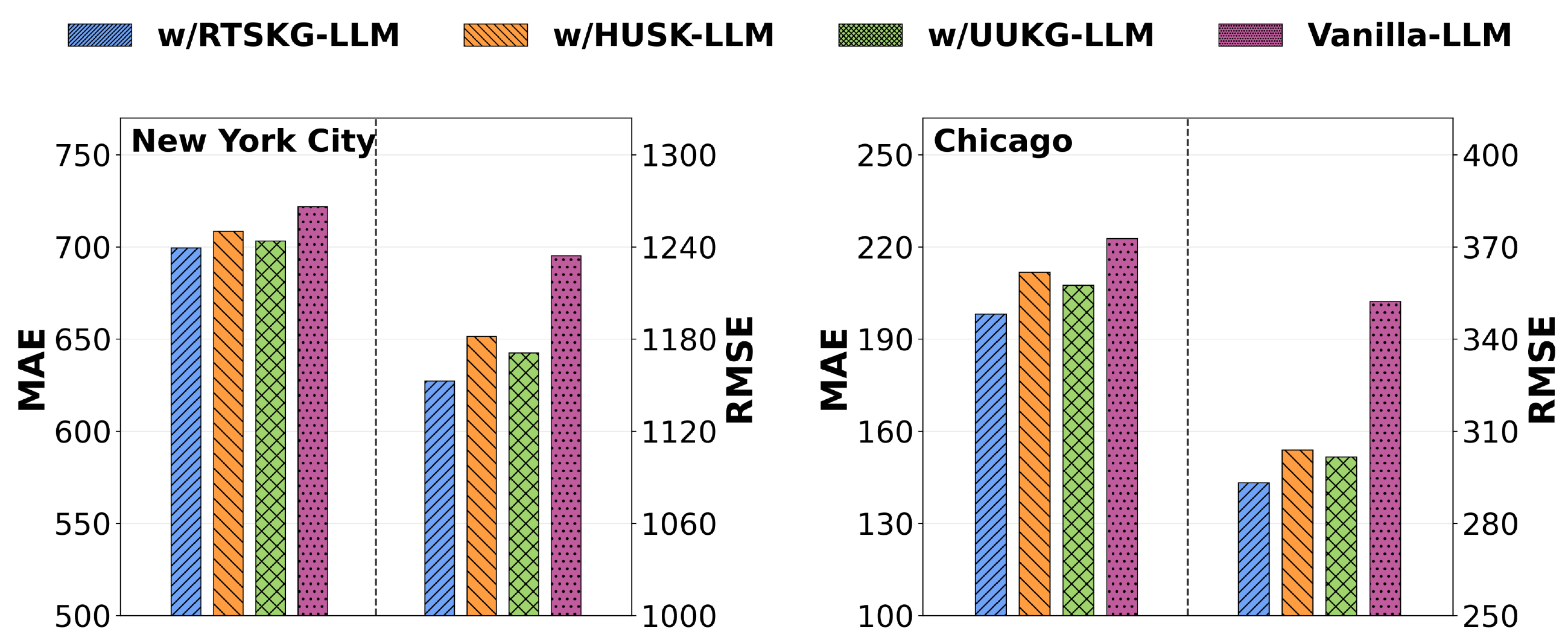}
  \caption{The comparison results of LLM-based knowledge-enhanced ridership prediction.}

\label{LLM}

\end{figure}

\paragraph{LLM-Based Knowledge-Enhanced Ridership Prediction.} Figure \ref{LLM} illustrates the comparison results of LLM-based knowledge-enhanced ridership prediction. Experimental results show that our RTSKG, as an external knowledge base for knowledge enhancement, achieves better performance than directly applying LLMs for ridership prediction and consistently outperforms leveraging other KGs for knowledge enhancement. This demonstrates that RTSKG is a high-quality dataset for city-level rail transit station analysis, and it maintains effectiveness even in the rapidly evolving field of LLM applications. RTSKG holds significant potential for supporting urban developments. In the future, RTSKG can be used to support more LLM-based applications, such as city-level rail transit station related Q\&A.

\section{Conclusion}
In this paper, we propose RTSKG, a new rail transit station knowledge graph dataset, which is designed to explicitly model rail transit stations and the complex spatial and semantic interactions among different kinds of urban entities for city-level rail transit station related tasks. Experimental results demonstrate that leveraging RTSKG 
for station-area store recommendation and knowledge-enhanced ridership prediction has the best performance compared with leveraging other urban KGs. 
By providing a reusable dataset for city-level rail transit station related tasks, our study has potential applications not only in rail transit systems but also in other urban infrastructure systems, which contributes to enhancing urban development.

\subsubsection{Acknowledgments.}
This work is supported by the NSFC (Grant No. 52378009, 62376058), ZhiShan Young Scholar Program of Southeast University (Grant No. 2242026RCB0011), the Fundamental Research Funds for the Central Universities, the Southeast University Interdisciplinary Research Program for Young Scholars, and the Big Data Computing Center of Southeast University.

\paragraph*{Resource Availability Statement:}
All classes, relations and instances in RTSKG are
identified by permanent dereferenceable URIs in w3id\footnote{\url{https://w3id.org/rtskg}}. All data are available as RDF dump files on Zenodo\footnote{\url{https://zenodo.org/records/21778005}}, and the basic information and source code of the RTSKG project
can be accessed at GitHub\footnote{\url{https://github.com/seucoin/RTSKG}}. The data resources are released under the CC BY-SA 4.0 License, and the source code is released under the Apache License 2.0.

\section*{Declaration of Use of Generative AI}
During the preparation of this work, generative AI tools were used exclusively for providing ideas of creating illustrative figures. The authors reviewed and approved all figures and take full responsibility for the content of the publication.

\bibliographystyle{splncs04}
\bibliography{iswc26}

\end{document}